# Probabilistic Learning of Torque Controllers from Kinematic and Force Constraints

João Silvério[1], Yanlong Huang[1], Leonel Rozo[1], Sylvain Calinon[2,1] and Darwin G. Caldwell[1]

*Abstract*— When learning skills from demonstrations, one is often required to think in advance about the appropriate task representation (usually in either operational or configuration space). We here propose a probabilistic approach for simultaneously learning and synthesizing torque control commands which take into account task space, joint space and force constraints. We treat the problem by considering different torque controllers acting on the robot, whose relevance is learned probabilistically from demonstrations. This information is used to combine the controllers by exploiting the properties of Gaussian distributions, generating new torque commands that satisfy the important features of the task. We validate the approach in two experimental scenarios using 7-DoF torque-controlled manipulators, with tasks that require the consideration of different controllers to be properly executed.

## I. INTRODUCTION

The field of Learning from Demonstration (LfD) [1] aims for a user-friendly and intuitive human-robot skill transfer. However, in general, when modeling demonstrations one must think in advance about the relevant variables to encode. Selecting these variables strongly depends on the task requirements, with motor skills often being represented in either operational or configuration space. The prior definition of the relevant space may require considerable reasoning or trial-and-error, contradicting the LfD concept. This process becomes even more cumbersome when the robot is required to physically interact with the environment, introducing additional task constraints such as interaction forces (the term *constraints* here refers to consistent features in demonstrations, that should be accurately reproduced). Consider the example shown in Fig. 1, where a robot is first required to apply a force with the end-effector, and then perform a configuration space movement. In this case, encoding demonstrations in either operational or configuration spaces alone will not result in proper execution.

We here propose an approach for simultaneously learning different types of task constraints and generating torque control commands that encapsulate the important features of the task. Figure 2 gives an overview of the approach. We treat the problem by considering different torque controllers acting on the robot, with each one being responsible for the fulfillment of a particular type of constraint (e.g. desired interaction forces, Cartesian/joint positions and/or velocities). We discuss such controllers in Section III. From demonstrations of a task, we propose to learn the importance of each controller using probabilistic representations of the collected data (Section V). We then exploit a set of linear operators, defined for each individual controller, that take into account the state of the robot and contact with the environment to transform the control references into torque commands, with associated importance. Finally, we combine the commands, represented as independent Gaussian-distributed torque references, through a fusion of controllers, carried out by a product of Gaussians (Section IV). We hence obtain a final torque reference, used to control the robot. Our contribution with respect to the state-of-the-art is three-fold:

1) A probabilistic formulation for jointly learning torque controllers from demonstrations, by exploiting the properties of Gaussian distributions.
2) The consideration of not only kinematic tasks (at Cartesian/joint space level) but also force-based ones.
3) An approach that is compatible with various probabilistic learning algorithms that generate Gaussian distributed references or trajectories.

[1] Department of Advanced Robotics, Istituto Italiano di Tecnologia, Genova, Italy (e-mail: name.surname@iit.it).
[2] Idiap Research Institute, Martigny, Switzerland (e-mail: sylvain.calinon@idiap.ch).
This work was supported by the Italian Ministry of Defense.

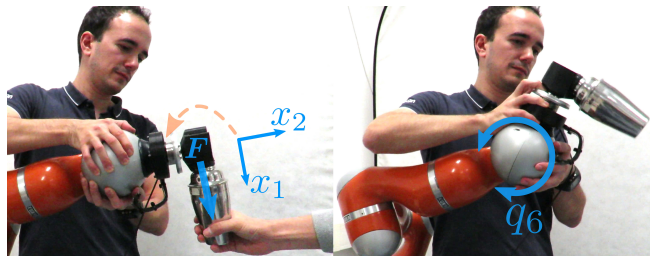

Fig. 1: Example of a task that demands two different controllers. First, the robot should close a shaker (left), by applying a force along $x_1$, a skill that requires force control. Subsequently, it must perform a shake with its wrist joint $q_6$ (right), thus a configuration space controller is desirable.

The proposed approach is evaluated in two scenarios with 7-DoF robots (Section VI). In the first case, we use a cocktail shaking task, employing force control, to prove that the approach can accommodate both force- and position/velocity-based skills. The second scenario shows that the approach can be used to combine partial demonstrations, allowing for demonstrating the sub-tasks of each controller independently.

## II. RELATED WORK

The problem of combining controllers can be broadly divided into two types of approaches. In [2], [3], [4], the authors use a weighted combination of individual torque controllers, with each controller responsible for a particular sub-task (e.g. balance, manipulation, joint limit avoidance). Other works frame the problem as a multi-level prioritization [5], [6], where lower importance tasks are executed without compromising more important ones, typically in a hierarchi-cal manner with a null space formulation. As a result, tasks with low importance are only executed if they do not affect high priority ones, potentially requiring platforms with a high number of degrees of freedom. Both kinds of approaches have their own merits, with the former allowing for a more flexible organization of tasks as well as smooth transitions between them (according to their weight profiles) and the latter ensuring that high priority tasks are always executed.

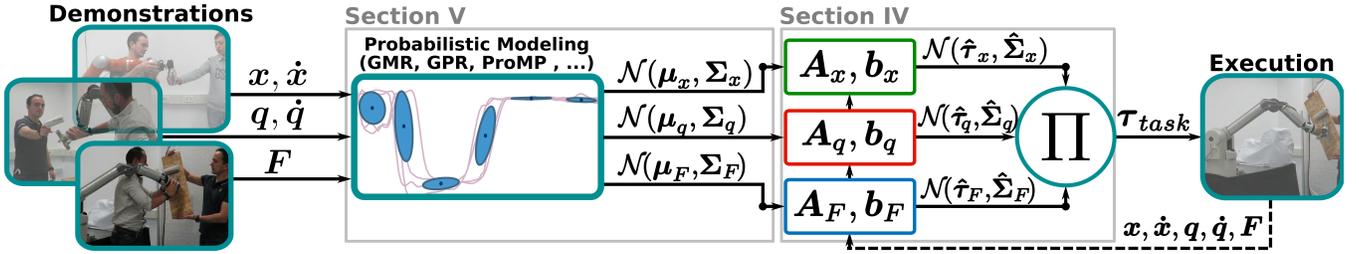

Fig. 2: Diagram of the proposed approach. Demonstrations of a task are given to the robot, while recording different types of data, such as positions, velocities and interaction forces. To each type of data, an individual controller is assigned, and the corresponding references are modeled as Gaussian distributions, encapsulating each controller's importance. During task execution, linear operators $\boldsymbol{A}$ and $\boldsymbol{b}$, which depend on the chosen controllers as well as the robot's state and the interaction forces, transform the references into probabilistic torque commands. These torques are combined by taking into account their variance, through the product of Gaussians, whose result is then fed to the robot as a torque $\boldsymbol{\tau}_{\text{task}}$ that satisfies the important task features.

In contrast to manually setting weights [2], in this paper we are interested in learning them from human demonstrations. Learning controller importance has been addressed in different manners, from reinforcement learning (RL) [3], [4], [7] to LfD [8], [9], [10]. The main differences between these two branches lie on the type of prior knowledge, with RL requiring *a priori* information in the form of reward or cost functions – which can be hard to formulate in some cases – and LfD approaches demanding task demonstrations. The present work shares connections with [8], [9], [10], where the problem of combining constraints in task and joint spaces is addressed. The first important difference is that such approaches use velocity controllers, which only take into account kinematic constraints. In this work, we exploit torque controllers, that allow for a straightforward consideration of desired interaction forces at the end-effector. Previous work in LfD has addressed learning forces, either alone [11] or in hybrid position-force control settings [12], [13], [14]. Here, we go one step further and consider interaction forces, Cartesian positions and joint trajectories simultaneously into the learning framework. The second relevant difference is that [8], [9], [10] model data using Gaussian Mixture Models (GMM), while in the present work, although GMM are used as an example, we generalize the solution to a wider range of probabilistic modeling approaches. In particular, we show that the probabilistic combination of controllers is compatible with any trajectory modeling technique that generates Gaussian-distributed outputs. Despite that we here showcase this property by exploiting Gaussian Process Regression (GPR) [15], in Section VI-B, other techniques such as Probabilistic Movement Primitives (ProMP) [16] may potentially be used.

### III. TORQUE CONTROLLERS – CONFIGURATION AND OPERATIONAL SPACE

Inspired by works in which a combination of torque controllers results in a flexible importance assignment and smooth transitions between different tasks [2], [3], [4], we propose a strategy where the controller combination is learned from demonstrations. In this section we define the individual controllers that we exploit for configuration and operational space control. Formally, we follow a model-based approach to control the robot using torques, by assuming a rigid-body system with $N$ joints whose dynamics are given by $\boldsymbol{M}(\boldsymbol{q})\ddot{\boldsymbol{q}} + \boldsymbol{C}(\boldsymbol{q},\dot{\boldsymbol{q}})\dot{\boldsymbol{q}} + \boldsymbol{g}(\boldsymbol{q}) = \boldsymbol{\tau}$, where $\boldsymbol{q}, \dot{\boldsymbol{q}}, \ddot{\boldsymbol{q}} \in \mathbb{R}^N$ denote joint angles, velocities and accelerations, $\boldsymbol{M}(\boldsymbol{q}) \in \mathbb{R}^{N\times N}$ corresponds to the inertia matrix, $\boldsymbol{C}(\boldsymbol{q},\dot{\boldsymbol{q}}) \in \mathbb{R}^N$ denotes the the Coriolis and centrifugal terms and $\boldsymbol{g}(\boldsymbol{q}) \in \mathbb{R}^N$ is a gravity term. The total torques acting on each joint are given by $\boldsymbol{\tau} \in \mathbb{R}^N$.

Robot control is achieved using a torque command $\boldsymbol{\tau}_u$, formed from a task-related term $\boldsymbol{\tau}_{\text{task}}$ and a term that compensates for the dynamics of the robot $\boldsymbol{\tau}_{\text{dyn}}$, i.e.,

$$\boldsymbol{\tau}_u = \boldsymbol{\tau}_{\text{task}} + \boldsymbol{\tau}_{\text{dyn}}, \quad (1)$$

where $\boldsymbol{\tau}_{\text{dyn}}$ is computed from the inverse dynamics model (assumed to be known). In this work we are interested in fusing controllers that fulfill different task requirements, thus we focus on the term $\boldsymbol{\tau}_{\text{task}}$. Here, when referring to tasks, we are concerned with the tracking of reference trajectories (e.g. positions, forces).

The definition of $\boldsymbol{\tau}_{\text{task}}$ depends on the space where tasks are represented. For instance, when a task requires the manipulation of an object (e.g. pick and place), $\boldsymbol{\tau}_{\text{task}}$ must be defined such that position and orientation constraints at the end-effector are fulfilled with respect to the object. If, additionally, manipulation requires physical contact (e.g. object insertion, cooperative handling of objects), $\boldsymbol{\tau}_{\text{task}}$ must also accommodate desired interaction forces. In other applications, where gestures or specific configurations of the kinematic chain are required, $\boldsymbol{\tau}_{\text{task}}$ is more adequately formulated as a configuration space controller. We now describe the controllers that we exploit for the different types of tasks, denoting $\boldsymbol{\tau}_{\text{task}}$ simply by $\boldsymbol{\tau}$.

#### A. Configuration space controller

Configuration space controllers are used to track joint positions and velocities. Here we exploit proportional-derivative (PD) controllers of the form

$$\boldsymbol{\tau}_q = \boldsymbol{K}_q^{\mathcal{P}}(\boldsymbol{q}_d - \boldsymbol{q}) + \boldsymbol{K}_q^{\mathcal{D}}(\dot{\boldsymbol{q}}_d - \dot{\boldsymbol{q}}), \quad (2)$$

where $\boldsymbol{K}_q^{\mathcal{P}}, \boldsymbol{K}_q^{\mathcal{D}} \in \mathbb{R}^{N \times N}$ are joint stiffness and damping gain matrices, and $\boldsymbol{q}, \dot{\boldsymbol{q}}, \boldsymbol{q}_d, \dot{\boldsymbol{q}}_d \in \mathbb{R}^N$ are the current and desired joint positions and velocities. An additional feed-forward term $\ddot{\boldsymbol{q}}_d \in \mathbb{R}^N$ is often added to (2), for improved tracking performance, as in [17]. As we shall see, it is straightforward to accommodate this term in our approach, if required.

### B. Position controller in operational space

Operational space controllers are aimed at tracking Cartesian poses with the end-effector of the robot. Here, we consider the case of tracking position references, but the approach remains valid for the consideration of orientations. We assume that the end-effector of the robot is driven by a force, that is proportional to the output of a PD controller,

$$\boldsymbol{F}_x = \bar{\boldsymbol{M}}(\boldsymbol{q})\Big(\boldsymbol{K}_x^{\mathcal{P}}(\boldsymbol{x}_d - \boldsymbol{x}) + \boldsymbol{K}_x^{\mathcal{D}}(\dot{\boldsymbol{x}}_d - \dot{\boldsymbol{x}})\Big), \quad (3)$$

where $\bar{\boldsymbol{M}}(\boldsymbol{q}) = (\boldsymbol{J}(\boldsymbol{q})\boldsymbol{M}(\boldsymbol{q})^{-1}\boldsymbol{J}(\boldsymbol{q})^\top)^{-1}$ is the Cartesian inertia matrix of the end-effector, whose positions and linear velocities (current and desired) are respectively denoted by $\boldsymbol{x}, \boldsymbol{x}_d, \dot{\boldsymbol{x}}, \dot{\boldsymbol{x}}_d \in \mathbb{R}^M$ (with $M$ being the dimension of the operational space). The Jacobian matrix $\boldsymbol{J}(\boldsymbol{q}) \in \mathbb{R}^{M \times N}$, gives the differential kinematics of the robot's end-effector $\dot{\boldsymbol{x}} = \boldsymbol{J}(\boldsymbol{q})\dot{\boldsymbol{q}}$ and $\boldsymbol{K}_x^{\mathcal{P}}, \boldsymbol{K}_x^{\mathcal{D}} \in \mathbb{R}^{M \times M}$ are Cartesian stiffness and damping gain matrices. The end-effector force $\boldsymbol{F}_x$ is converted to joint torques as in [17],

$$\boldsymbol{\tau}_x = \boldsymbol{J}(\boldsymbol{q})^\top \boldsymbol{F}_x. \quad (4)$$

### C. Force controller

In this case we consider a proportional controller that tracks a desired force at the end-effector (see [18], Ch. 11):

$$\boldsymbol{F}_u = \boldsymbol{K}_F^{\mathcal{P}}(\boldsymbol{F}_d - \boldsymbol{F}), \qquad \boldsymbol{\tau}_F = \boldsymbol{J}(\boldsymbol{q})^\top \boldsymbol{F}_u, \quad (5)$$

where $\boldsymbol{F}, \boldsymbol{F}_d \in \mathbb{R}^M$ are current and desired contact forces (measured using a F/T sensor at the end-effector), and (4) is used to map the force command at the end-effector to joint torques*. Finally, $\boldsymbol{K}_F^{\mathcal{P}} \in \mathbb{R}^{M \times M}$ is a proportional gain matrix.

## IV. PROBABILISTIC TORQUE CONTROLLERS

In this section, we formalize the fusion of torque controllers as an optimization problem and lay out the probabilistic treatment of control commands. Let us consider a robot employing $P$ controllers – as those defined in Section III – at any given moment, corresponding to $P$ different sub-tasks that can be executed in series or in parallel. Each controller generates a torque command $\boldsymbol{\tau}^{(p)} \in \mathbb{R}^N$, $p = 1, \ldots, P$. Also, let us assume we have access to a precision matrix (which will be explained in Section IV-B), denoted by $\boldsymbol{\Gamma}^{(p)} \in \mathbb{R}^{N \times N}$, providing information about the respective importance of the different controllers. We formalize the problem of fusing $P$ control commands as the optimization

$$\hat{\boldsymbol{\tau}} = \arg\min_{\boldsymbol{\tau}} \sum_{p=1}^{P} \big(\boldsymbol{\tau} - \boldsymbol{\tau}^{(p)}\big)^\top \boldsymbol{\Gamma}^{(p)} \big(\boldsymbol{\tau} - \boldsymbol{\tau}^{(p)}\big), \quad (6)$$

*In the remainder of the paper we drop dependencies on $\boldsymbol{q}$, e.g. $\bar{\boldsymbol{M}} = \bar{\boldsymbol{M}}(\boldsymbol{q}), \boldsymbol{J} = \boldsymbol{J}(\boldsymbol{q})$, etc.

whose objective function corresponds to a weighted sum of quadratic error terms, with the weight of each term given by the matrices $\boldsymbol{\Gamma}^{(p)}$. The solution and error residuals of (6) can be computed analytically, and correspond to the mean and covariance matrix of a Gaussian distribution $\mathcal{N}(\hat{\boldsymbol{\tau}}, \hat{\boldsymbol{\Sigma}}_\tau)$ given by the product of $P$ Gaussians, with means $\boldsymbol{\tau}^{(p)}$ and precision matrices $\boldsymbol{\Gamma}^{(p)}$,

$$\hat{\boldsymbol{\tau}} = \hat{\boldsymbol{\Sigma}}_\tau \sum_{p=1}^P \boldsymbol{\Gamma}^{(p)} \boldsymbol{\tau}^{(p)}, \qquad \hat{\boldsymbol{\Sigma}}_\tau = \Big(\sum_{p=1}^P \boldsymbol{\Gamma}^{(p)}\Big)^{-1}, \quad (7)$$

where precision matrices are the inverse of covariance matrices $\boldsymbol{\Sigma}_\tau^{(p)}$, i.e. $\boldsymbol{\Gamma}^{(p)} = \boldsymbol{\Sigma}_\tau^{(p)^{-1}}$. The connection between the solution of (6) and the product of Gaussians (7) allows for exploiting the structure of the controllers defined in Section III to fuse torque commands, given Gaussian-distributed references. In particular, this is achieved by taking advantage of the linearity of the controllers (Section IV-A) in combination with the linear properties of Gaussians (Section IV-B).

### A. Linear controller structure

Control commands (2)–(5) are linear with respect to the reference trajectories. The controller equations can thus be re-written in a way that highlights this linear structure. For the joint space torque controller (2) we obtain

$$\boldsymbol{\tau}_q = \begin{bmatrix} \boldsymbol{K}_q^{\mathcal{P}} & \boldsymbol{K}_q^{\mathcal{D}} \end{bmatrix} \begin{bmatrix} \boldsymbol{q}_d \\ \dot{\boldsymbol{q}}_d \end{bmatrix} - \begin{bmatrix} \boldsymbol{K}_q^{\mathcal{P}} & \boldsymbol{K}_q^{\mathcal{D}} \end{bmatrix} \begin{bmatrix} \boldsymbol{q} \\ \dot{\boldsymbol{q}} \end{bmatrix}$$

$$\Leftrightarrow \boldsymbol{\tau}_q = \boldsymbol{A}_q \begin{bmatrix} \boldsymbol{q}_d \\ \dot{\boldsymbol{q}}_d \end{bmatrix} + \boldsymbol{b}_q, \quad (8)$$

where $\boldsymbol{A}_q = \begin{bmatrix} \boldsymbol{K}_q^{\mathcal{P}} & \boldsymbol{K}_q^{\mathcal{D}} \end{bmatrix}$ and $\boldsymbol{b}_q = -\begin{bmatrix} \boldsymbol{K}_q^{\mathcal{P}} & \boldsymbol{K}_q^{\mathcal{D}} \end{bmatrix} \begin{bmatrix} \boldsymbol{q} \\ \dot{\boldsymbol{q}} \end{bmatrix}$. Similarly, the Cartesian position and force controllers (4)–(5) can be formulated as $\boldsymbol{\tau}_x = \boldsymbol{A}_x \begin{bmatrix} \boldsymbol{x}_d \\ \dot{\boldsymbol{x}}_d \end{bmatrix} + \boldsymbol{b}_x$, with $\boldsymbol{A}_x = \boldsymbol{J}^\top \bar{\boldsymbol{M}} \begin{bmatrix} \boldsymbol{K}_x^{\mathcal{P}} & \boldsymbol{K}_x^{\mathcal{D}} \end{bmatrix}$, $\boldsymbol{b}_x = -\boldsymbol{J}^\top \bar{\boldsymbol{M}} \begin{bmatrix} \boldsymbol{K}_x^{\mathcal{P}} & \boldsymbol{K}_x^{\mathcal{D}} \end{bmatrix} \begin{bmatrix} \boldsymbol{x} \\ \dot{\boldsymbol{x}} \end{bmatrix}$, and $\boldsymbol{\tau}_F = \boldsymbol{A}_F \boldsymbol{F}_d + \boldsymbol{b}_F$, with $\boldsymbol{A}_F = \boldsymbol{J}^\top \boldsymbol{K}_F^{\mathcal{P}}$ and $\boldsymbol{b}_F = -\boldsymbol{J}^\top \boldsymbol{K}_F^{\mathcal{P}} \boldsymbol{F}$. Note that linearity also applies if feed-forward terms are included in the controllers, e.g. $\ddot{\boldsymbol{x}}_d, \ddot{\boldsymbol{q}}_d$. In such cases, these terms simply need to be included in the reference vector and $\boldsymbol{A}$ can be extended with the identity matrix, e.g. $[\boldsymbol{q}_d^\top \ \dot{\boldsymbol{q}}_d^\top \ \ddot{\boldsymbol{q}}_d^\top]^\top$ and $\boldsymbol{A}_q = \begin{bmatrix} \boldsymbol{K}_q^{\mathcal{P}} & \boldsymbol{K}_q^{\mathcal{D}} & \boldsymbol{I} \end{bmatrix}$, for a configuration space controller.

### B. From probabilistic references to probabilistic torques

Gaussian distributions are popular in robot learning and control due to their properties of product, conditioning and linear transformation. Here, we consider Gaussian-distributed control references and exploit the previously defined linear operators to formulate probabilistic torque controllers. Let us first consider a configuration space controller, with desired joint state $\begin{bmatrix} \boldsymbol{q}_d \\ \dot{\boldsymbol{q}}_d \end{bmatrix} \sim \mathcal{N}\big(\boldsymbol{\mu}_q, \boldsymbol{\Sigma}_q\big)$, where $\boldsymbol{\mu}_q \in \mathbb{R}^{2N}$ and $\boldsymbol{\Sigma}_q \in \mathbb{R}^{2N \times 2N}$ are the mean and covariance matrix of a Gaussian, modeling the probability distribution of joint positions and velocities. Per the linear properties of Gaussian distributions, the configuration space controller (8) yields a new Gaussian $\mathcal{N}(\boldsymbol{\tau}_q, \boldsymbol{\Sigma}_{\tau,q})$ with mean and covariance given by

$$\boldsymbol{\tau}_q = \boldsymbol{A}_q \boldsymbol{\mu}_q + \boldsymbol{b}_q, \quad \boldsymbol{\Sigma}_{\tau,q} = \boldsymbol{A}_q \boldsymbol{\Sigma}_q \boldsymbol{A}_q^\top. \quad (9)$$

Similarly, for $\begin{bmatrix} \boldsymbol{x}_d \\ \dot{\boldsymbol{x}}_d \end{bmatrix} \sim \mathcal{N}(\boldsymbol{\mu}_x, \boldsymbol{\Sigma}_x)$ and $\boldsymbol{F}_d \sim \mathcal{N}(\boldsymbol{\mu}_F, \boldsymbol{\Sigma}_F)$, we obtain

$$\boldsymbol{\tau}_x = \boldsymbol{A}_x \boldsymbol{\mu}_x + \boldsymbol{b}_x, \quad \boldsymbol{\Sigma}_{\tau,x} = \boldsymbol{A}_x \boldsymbol{\Sigma}_x \boldsymbol{A}_x^\top, \quad (10)$$

and

$$\boldsymbol{\tau}_F = \boldsymbol{A}_F \boldsymbol{\mu}_F + \boldsymbol{b}_F, \quad \boldsymbol{\Sigma}_{\tau,F} = \boldsymbol{A}_F \boldsymbol{\Sigma}_F \boldsymbol{A}_F^\top, \quad (11)$$

respectively. This type of controller has a probabilistic nature as the torque commands are generated from Gaussian distributions and result in new Gaussians. We therefore refer to them as *probabilistic torque controllers* (PTC).

A generic PTC, $p = 1, \ldots, P$, is thus fully specified by

$$\boldsymbol{\xi}_d \sim \mathcal{N}(\boldsymbol{\mu}^{(p)}, \boldsymbol{\Sigma}^{(p)}), \quad \{\boldsymbol{A}^{(p)}, \boldsymbol{b}^{(p)}\}, \quad (12)$$
$$\boldsymbol{\tau}^{(p)} = \boldsymbol{A}^{(p)} \boldsymbol{\mu}^{(p)} + \boldsymbol{b}^{(p)}, \quad \boldsymbol{\Sigma}_\tau^{(p)} = \boldsymbol{A}^{(p)} \boldsymbol{\Sigma}^{(p)} \boldsymbol{A}^{(p)\top},$$

where $\boldsymbol{\xi}_d$ denotes a generic control reference. Note that the set of linear parameters $\{\boldsymbol{A}^{(p)}, \boldsymbol{b}^{(p)}\}$ is permanently updated, for each controller, during execution, as it depends on the state of the robot and its interaction with the environment through $\boldsymbol{q}, \dot{\boldsymbol{q}}, \boldsymbol{x}, \dot{\boldsymbol{x}}$ and $\boldsymbol{F}$.

A probabilistic representation of trajectories using Gaussian distributions (12) has the advantage of modeling the second moment of the data in the form of covariance matrices. This is exploited here to express the importance of each controller – denoted by $\boldsymbol{\Gamma}^{(p)}$ – as a function of the covariance of the corresponding reference trajectory $\boldsymbol{\Sigma}^{(p)}$:

$$\boldsymbol{\Gamma}^{(p)} = \boldsymbol{\Sigma}_\tau^{(p)^{-1}} = \left(\boldsymbol{A}^{(p)} \boldsymbol{\Sigma}^{(p)} \boldsymbol{A}^{(p)\top}\right)^{-1}. \quad (13)$$

Note that $\boldsymbol{A}^{(p)}$ is typically non-squared. This operator maps constraints from spaces with different dimensions (e.g. configuration and operational spaces) into a common space, that of torque commands.

With the definition of $\boldsymbol{\Gamma}^{(p)}$ in (13), torque commands can be combined using (7). The problem of learning control commands and their respective importance is thus framed as the learning of reference trajectories as Gaussian distributions $\mathcal{N}(\boldsymbol{\mu}^{(p)}, \boldsymbol{\Sigma}^{(p)})$, and generating Gaussian-distributed torque commands $\mathcal{N}(\boldsymbol{\tau}^{(p)}, \boldsymbol{\Sigma}_\tau^{(p)})$, which encapsulate the control reference and its importance with respect to other controllers. In previous work, controller weights are either set empirically [2] or optimized through reinforcement learning [3], [4]. In contrast to these works, we employ probabilistic regression algorithms to learn $\mathcal{N}(\boldsymbol{\mu}^{(p)}, \boldsymbol{\Sigma}^{(p)})$, and consequently $\mathcal{N}(\boldsymbol{\tau}^{(p)}, \boldsymbol{\Sigma}_\tau^{(p)}), \forall p = 1, ..., P$, from human demonstrations.

## V. Learning control references from demonstrations

In Section IV, we formalized our approach for combining controllers. Here we show how the Gaussian modeling of trajectories can be learned from demonstrations. Several regression methods exist for this purpose, each offering different advantages; see [19] for a review. Two popular approaches are GMM, combined with Gaussian Mixture Regression [20], and GPR [15]. We now review these two techniques, and expand on their use in the context of PTC.

### A. Gaussian Mixture Model/Gaussian Mixture Regression (GMM/GMR)

We consider demonstration datasets comprised of $T$ datapoints organized in a matrix $\boldsymbol{\xi} \in \mathbb{R}^{D \times T}$. Each datapoint $\boldsymbol{\xi}_t$ is represented with input/output dimensions indexed by $\mathcal{I}$, $\mathcal{O}$, so that $\boldsymbol{\xi}_t = \begin{bmatrix} \boldsymbol{\xi}_t^\mathcal{I} \\ \boldsymbol{\xi}_t^\mathcal{O} \end{bmatrix} \in \mathbb{R}^D$ with $D = D_\mathcal{I} + D_\mathcal{O}$. It can for example represent a concatenation of time stamps with end-effector poses, joint angles or measured forces. A GMM, encoding the joint probability distribution $\mathcal{P}(\boldsymbol{\xi}^\mathcal{I}, \boldsymbol{\xi}^\mathcal{O})$ with $K$ states and parameters $\boldsymbol{\Theta} = \{\pi_i, \boldsymbol{\mu}_i, \boldsymbol{\Sigma}_i\}_{i=1}^K$ (respectively the prior, mean and covariance matrix of each state $i$), can be estimated from such a dataset through Expectation-Maximization (EM) [20]. After a GMM is fitted to a given dataset, GMR can subsequently be used to synthesize new behaviors, for new inputs $\boldsymbol{\xi}_*^\mathcal{I} \in \mathbb{R}^{D_\mathcal{I}}$, by means of the conditional probability $\mathcal{P}(\boldsymbol{\xi}_*^\mathcal{O} | \boldsymbol{\xi}_*^\mathcal{I})$, yielding a normally-distributed output $\boldsymbol{\xi}_*^\mathcal{O} | \boldsymbol{\xi}_*^\mathcal{I} \sim \mathcal{N}(\boldsymbol{\mu}_\mathcal{O}, \boldsymbol{\Sigma}_\mathcal{O})$; see [20] for details.

We exploit GMM/GMR to estimate desired trajectories for each controller through the mean $\boldsymbol{\mu}_\mathcal{O}$, as well as their importance through the covariance matrix $\boldsymbol{\Sigma}_\mathcal{O}$. In GMM/GMR, covariance matrices model the variability in the data, in addition to the correlation between the variables. Figure 3(a) illustrates this aspect, where we see that the variance regressed by GMR (shown as an envelope around the mean) reflects the datapoint distribution in the original dataset. In the context of PTCs, high variability in the demonstrations of the $p$-th controller results in large covariance matrices $\boldsymbol{\Sigma}^{(p)}$. From (13), it follows that the corresponding controller precision matrix $\boldsymbol{\Gamma}^{(p)}$ will be small and, thus, the control reference $\boldsymbol{\tau}^{(p)}$ will be tracked less accurately. GMM/GMR is, hence, an appropriate technique to select relevant controllers based on the regularities observed in each part of the task throughout the different demonstrations.

### B. Gaussian Process Regression (GPR)

A Gaussian Process (GP) is a distribution over functions, with a Gaussian prior on observations $\boldsymbol{\xi}^\mathcal{O}$ given by $\boldsymbol{\xi}^\mathcal{O} \sim \mathcal{N}(\boldsymbol{m}(\boldsymbol{\Xi}^\mathcal{I}), \boldsymbol{K}(\boldsymbol{\Xi}^\mathcal{I}, \boldsymbol{\Xi}^\mathcal{I}))$, where $\boldsymbol{m}(\boldsymbol{\Xi}^\mathcal{I})$ is a vector-valued function yielding the mean of the process, $\boldsymbol{K}(\boldsymbol{\Xi}^\mathcal{I}, \boldsymbol{\Xi}^\mathcal{I})$ denotes its covariance matrix and $\boldsymbol{\Xi}^\mathcal{I} = [\boldsymbol{\xi}_1^\mathcal{I} \ldots \boldsymbol{\xi}_T^\mathcal{I}] \in \mathbb{R}^{D_\mathcal{I} \times T}$ is a concatenation of observed inputs. The covariance matrix is computed from a kernel function evaluated at the inputs, with elements $K_{ij} = k(\boldsymbol{\xi}_i^\mathcal{I}, \boldsymbol{\xi}_j^\mathcal{I})$. Several types of kernel functions exist; see e.g., [15].

Standard GPR allows for predicting a scalar function $\boldsymbol{\xi}_*^\mathcal{O} = f(\boldsymbol{\xi}_*^\mathcal{I}) : \mathbb{R}^{D_\mathcal{I}} \to \mathbb{R}$. In robotics, one typically requires multi-dimensional outputs, thus GPR is often employed separately for each output of a given problem. Here we follow this approach to probabilistically model multi-dimensional reference trajectories, such as those of joint angles or Cartesian positions. For each input point $\boldsymbol{\xi}_*^\mathcal{I} \in \mathbb{R}^{D_\mathcal{I}}$, the prediction

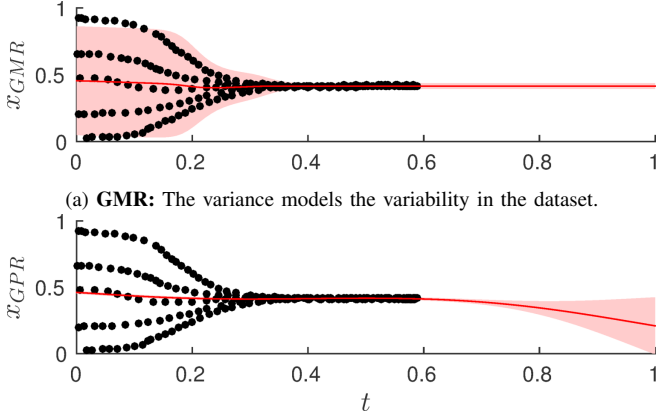

(a) **GMR:** The variance models the variability in the dataset.

(b) **GPR:** The variance models the uncertainty of the estimate (depending on the presence/absence of training datapoints in the neighborhood).

Fig. 3: For a given set of datapoints (black dots), GMR and GPR compute different and complementary notions of variance. The red line is the regressed function, while the light red contour represents the computed variance around the prediction.

---

**Algorithm 1** Fusion of probabilistic torque controllers

  ***1. Initialization***
1: Select $P$ controllers (Section III) based on the task
2: Select appropriate regression algorithm (GMR, GPR)
3: Collect demonstrations for each controller $\{\boldsymbol{\xi}_p^{\mathcal{I}}, \boldsymbol{\xi}_p^{\mathcal{O}}\}_{p=1}^{P}$
  ***2. Model training***
1: **for** $p = 1, \ldots, P$ **do**
2:    **if** regression algorithm is GMR **then**
3:       Choose GMM state number $K$ and estimate $\Theta$
4:    **else if** regression algorithm is GPR **then**
5:       Choose the kernel $k(\cdot, \cdot)$ and its hyperparameters
6:    **end if**
7: **end for**
  ***3. Movement synthesis***
1: **for** $t = 1, \ldots, T$ **do**
2:    **for** $p = 1, \ldots, P$ **do**
3:       Compute $\boldsymbol{\mu}^{(p)}, \boldsymbol{\Sigma}^{(p)} | \boldsymbol{\xi}_t^{\mathcal{I}}$, through GMR or GPR
4:       Update $\{\boldsymbol{A}^{(p)}, \boldsymbol{b}^{(p)}\}$ based on the type of controller
5:       Compute torque distribution $\mathcal{N}\left(\boldsymbol{\tau}^{(p)}, \boldsymbol{\Sigma}_\tau^{(p)}\right)$
6:    **end for**
7:    Compute $\hat{\boldsymbol{\tau}}$ from (7) and $\boldsymbol{\tau}_u$ from (1)
8: **end for**

---

of each output dimension $d = 1, \ldots, D_{\mathcal{O}}$ is thus given by

$$\mu_d = \boldsymbol{m}_* + \boldsymbol{k}_*[\boldsymbol{K} + \epsilon_n^2 \boldsymbol{I}]^{-1}(\boldsymbol{\xi}^{\mathcal{O}_d} - \boldsymbol{m}), \quad (14)$$
$$\sigma_d^2 = \boldsymbol{k}_{**} - \boldsymbol{k}_*[\boldsymbol{K} + \epsilon_n^2 \boldsymbol{I}]^{-1}\boldsymbol{k}_*, \quad (15)$$

where $\boldsymbol{\xi}^{\mathcal{O}_d} \in \mathbb{R}^T$ is the observed $d$-th output dimension, $\boldsymbol{k}_* = [k(\boldsymbol{\xi}_*^{\mathcal{I}}, \boldsymbol{\xi}_1^{\mathcal{I}}) \ldots k(\boldsymbol{\xi}_*^{\mathcal{I}}, \boldsymbol{\xi}_T^{\mathcal{I}})]$, $\boldsymbol{k}_{**} = k(\boldsymbol{\xi}_*^{\mathcal{I}}, \boldsymbol{\xi}_*^{\mathcal{I}})$, $\boldsymbol{m} = \boldsymbol{m}(\boldsymbol{\Xi}^{\mathcal{I}})$, $\boldsymbol{m}_* = \boldsymbol{m}(\boldsymbol{\xi}_*^{\mathcal{I}})$, $\boldsymbol{K} = \boldsymbol{K}(\boldsymbol{\Xi}^{\mathcal{I}}, \boldsymbol{\Xi}^{\mathcal{I}})$, and $\epsilon_n^2$ is an additional hyperparameter modeling noise in the observations (which acts as a regularization term). We concatenate the predictions into one single multivariate Gaussian with mean and covariance matrix given by

$$\boldsymbol{\mu}_{\mathcal{O}} = \begin{bmatrix} \mu_1 & \ldots & \mu_{D_{\mathcal{O}}} \end{bmatrix}^\top, \quad \boldsymbol{\Sigma}_{\mathcal{O}} = \mathrm{diag}(\sigma_1^2, \ldots, \sigma_{D_{\mathcal{O}}}^2). \quad (16)$$

Since output dimensions are modeled separately, GPR predictions are, in the standard case, uncorrelated, which is evident from the structure of $\boldsymbol{\Sigma}_{\mathcal{O}}$ in (16). In contrast to GMR, the estimated variance in GPR is a measure of prediction uncertainty. Figure 3(b) illustrates this aspect, with the variance increasing with the absence of training data ($t > 0.6s$). This provides a way of assigning importance to predictions, when different observations of a task occur. We propose to exploit GPR if demonstration data is incomplete or scarce and, in particular, for partially demonstrating a task to each controller as separate sub-tasks.

The overall approach is summarized in Algorithm 1 for GMM or GP as trajectory modeling techniques.

## VI. EVALUATION

We assess the performance of the proposed framework in two different tasks. In one case, we exploit the variability in the demonstrations, while, in the other, we consider the prediction uncertainty. The experiments are conducted in two different 7-DoF manipulators, enabled with torque control. The reader is referred to http://joaosilverio.weebly.com/iros18 for videos of both experiments.

### A. Learning cocktail shaking skills with force constraints

We start our evaluation with a cocktail shaking task where force and configuration space control are employed. For this task we use the torque-controlled KUKA light-weight robot. The task is comprised of two sub-tasks (Fig. 1): a force-based sub-task, where an interaction force (measured with a F/T sensor mounted on the end-effector) must be tracked in order to successfully close a cocktail shaker, and a configuration space sub-task, through which the robot performs a shake using rhythmic joint movements. A joint space encoding of the shaking movement is more likely to generate a proper reproduction since rhythmic movements are typically less consistent in operational space than in joint space [9]. We aim to extract the activation of each sub-task from the variability in the demonstrations, thus both force and joint demonstrations are encoded in GMMs, together with time, which is used as the input to GMR.

We collected 4 demonstrations of this task by kinesthetically guiding the robot arm (gravity-compensated) to first close the shaker and, second, to perform the shake with a rhythmic motion of its 6th joint (see Fig. 1). For $p = 1$, the force controller, we have $D = 4$, with datapoints encoding time and sensed forces $\boldsymbol{\xi}_t^{(1)} = [\, t \; F_{1,t} \; F_{2,t} \; F_{3,t} \,]^\top$ (force directions as indicated in Fig. 1). In the case of the joint space controller, $p = 2$, we have $D = 15$ with datapoints $\boldsymbol{\xi}_t^{(2)} = [\, t \; q_{1,t} \; \ldots \; q_{7,t} \; \dot{q}_{1,t} \; \ldots \; \dot{q}_{7,t} \,]^\top$, where $q_{n,t}$ and $\dot{q}_{n,t}$ denote the position and velocity of joint $n$ at time step $t$. The recorded trajectories were filtered and sub-sampled to 200 points each, yielding a dataset with $T = 800$ datapoints for each controller. Additionally, the joint space trajectories were aligned using Dynamic Time Warping, in order to capture the consistent shaking patterns in all demonstrations. Finally, GMMs were fitted to the dataset of each controller, with $K = 7$ and $K = 15$ states, respectively, chosen empirically.

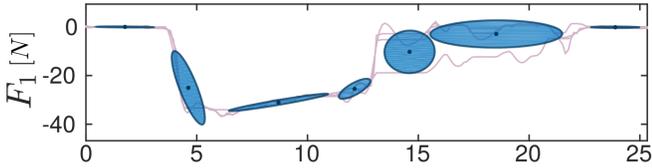

Fig. 4: Dataset of demonstrated contact forces along $F_1$ (lines) and estimated GMM states (blue ellipses).

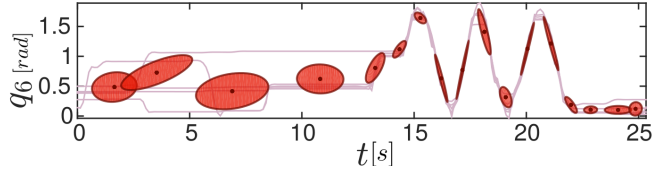

Fig. 5: Dataset from joint $q_6$ of the 7-DoF manipulator as a function of time (lines). Red ellipses are the GMM states which model the joint probability distribution between joint angles and time.

Figures 4 and 5 show the force and joint space datapoints, together with the corresponding GMM states, for $F_1$ (force along the end-effector $x_1$-axis) and joint $q_6$. For illustration purposes, the GMM states are depicted as ellipses with a width of one standard deviation. The negative sign in the force measurements indicates that the applied force is in opposite direction to the positive $x_1$-axis, which is expected due to the closing of the shaker occurring along that direction. From these plots we conclude that both the collected contact forces and joint angles have periods of high and low variability. The periods of low variability mark the regions where each sub-task should be predominant. In the case of $F_1$, this happens at the beginning, where the force is zero, and between $5s$ and $10s$, where the contact force is applied to close the shaker. On the other hand, the consistent rhythmic patterns after $t = 15s$ in Fig. 5, mark the shaking sub-task. Notably, in both cases, the GMM encoding is able to capture this consistency, in the form of narrow Gaussians. Figure 6 shows the retrieved control references using GMR, given the time input. Here, the contours around thick lines correspond to the predicted variance at each input point. In both cases, GMM/GMR allows for a proper encoding and retrieval of both mean control reference and variance.

The torque commands that were generated by each controller during one reproduction of the task, as well as the optimal torque, are shown in Fig. 7. The latter is obtained from the former two using (6), as described in Section IV. We focus our analysis on joint $q_6$, the one which performs the shake. For each sub-task, we used diagonal control gain matrices, chosen empirically based on the desired tracking precision. In particular, we used $\boldsymbol{K}_F^{\mathcal{P}} = \mathrm{diag}(4, 2, 2)$, $\boldsymbol{K}_q^{\mathcal{P}} = \mathrm{diag}(50, 80, 20, 70, 20, 10, 6)$ and $\boldsymbol{K}_q^{\mathcal{D}} = \mathrm{diag}(14, 17, 8, 16, 8, 6, 4)$. The linear operators $\{\boldsymbol{A}^{(1)}, \boldsymbol{b}^{(1)}\}, \{\boldsymbol{A}^{(2)}, \boldsymbol{b}^{(2)}\}$ were constructed according to Section IV-B as $\boldsymbol{A}_t^{(1)} = -\boldsymbol{J}_t^\top \boldsymbol{K}_F^{\mathcal{P}}$ and $\boldsymbol{b}_t^{(1)} = \boldsymbol{J}_t^\top \boldsymbol{K}_F^{\mathcal{P}} \boldsymbol{F}_t$, for the contact force controller, and $\boldsymbol{A}_t^{(2)} = \begin{bmatrix} \boldsymbol{K}_q^{\mathcal{P}} & \boldsymbol{K}_q^{\mathcal{D}} \end{bmatrix}$ and $\boldsymbol{b}_t^{(2)} = -\begin{bmatrix} \boldsymbol{K}_q^{\mathcal{P}} & \boldsymbol{K}_q^{\mathcal{D}} \end{bmatrix} \begin{bmatrix} \boldsymbol{q}_t \\ \dot{\boldsymbol{q}}_t \end{bmatrix}$, for the configuration space controller. Notice the sign change in the force operators, compared to those in Section IV-B. This is due to the encoded forces having an opposite sign to the desired direction of end-effector movement. Figure 7 shows that the commanded torque closely matches the torque from each of the individual controllers, in the corresponding regions of low variance (note that the weight of each controller is inversely proportional to the variance, as per Eq. (13)). This is evident in the beginning of the task, where the torques generated by the force controller strongly influence the torques sent to the robot, and from $t = 15s$, where the shaking torques are favored. This results in a reproduction where the complete task is properly executed by, first, applying the desired contact force and, second, performing the shaking movement.

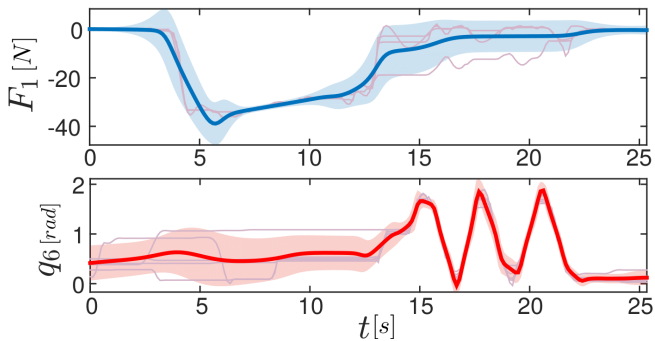

Fig. 6: GMR performed on the mixture models depicted in Figs. 4 and 5, with solid lines representing the retrieved profiles and the semi-transparent contours depicting the prediction variance. **Top:** Retrieved contact force profile $F_1$. **Bottom:** Predicted reference for $q_6$.

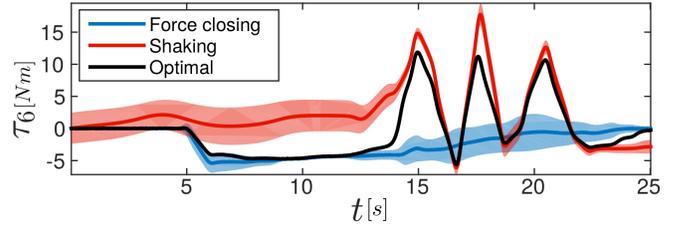

Fig. 7: Generated torque commands for joint 6 during one reproduction of the task. Red and blue curves show the torques generated by each individual controller, with corresponding variance, obtained from the probabilistic controller formulation in Section IV. The optimal torque, used by the robot, is depicted in black.

### B. Learning painting skills from separate demonstrations

In a second experiment we aim at showing that our framework is compatible with probabilistic techniques other than GMM. Here, we consider the scenario where a robot assists a user to perform a painting task. We divide the complete task into two sub-tasks: 1) a handover, where the user gives the paint roller to the robot (Fig. 8-*left*), and 2) painting, where, in a different region of the workspace, the robot helps the user paint a wooden board by applying painting strokes (Fig. 8-*right*). We employ an operational space controller (4) for the handover and a configuration space controller (2) for the painting.

Teaching controllers separately implies a trajectory modeling technique that yields high variances when far from each controller training region, thus we exploit GPR. The 3-dimensional position of the user right hand is, in this case, used as an input to GPR, as opposed to time. Training datapoints have the form $\boldsymbol{\xi}_t^{(1)} = [\boldsymbol{x}_t^H \ \boldsymbol{x}_t^R]$ for the handover sub-task and $\boldsymbol{\xi}_t^{(2)} = [\boldsymbol{x}_t^H \ \boldsymbol{q}_t]$ for the painting sub-task. Here, $\boldsymbol{x}_t^H, \boldsymbol{x}_t^R \in \mathbb{R}^3$ are the human and robot hand positions at time $t$ and $\boldsymbol{q}_t \in \mathbb{R}^7$ is the joint space configuration of the manipulator. The reference trajectories of each sub-task are thus 3- and 7-dimensional, respectively. In this experiment we consider zero velocity references for both controllers, $\dot{\boldsymbol{x}}_d = \boldsymbol{0}$, $\dot{\boldsymbol{q}}_d = \boldsymbol{0}$, and thus we used linear operators $\boldsymbol{A}_t^{(1)} = \boldsymbol{J}_t^\top \bar{\boldsymbol{M}}_t \boldsymbol{K}_x^\mathcal{P}$, $\boldsymbol{b}_t^{(1)} = -\boldsymbol{J}_t^\top \bar{\boldsymbol{M}}_t \begin{bmatrix} \boldsymbol{K}_x^\mathcal{P} & \boldsymbol{K}_x^\mathcal{D} \end{bmatrix} \begin{bmatrix} \boldsymbol{x}_t \\ \dot{\boldsymbol{x}}_t \end{bmatrix}$ and $\boldsymbol{A}_t^{(2)} = \boldsymbol{K}_q^\mathcal{P}$ and $\boldsymbol{b}_t^{(2)} = -\begin{bmatrix} \boldsymbol{K}_q^\mathcal{P} & \boldsymbol{K}_q^\mathcal{D} \end{bmatrix} \begin{bmatrix} \boldsymbol{q}_t \\ \dot{\boldsymbol{q}}_t \end{bmatrix}$. Moreover, we set $\boldsymbol{K}_x^\mathcal{P} = \mathrm{diag}(75, 75, 75)$, $\boldsymbol{K}_x^\mathcal{D} = \mathrm{diag}(17.5, 17.5, 17.5)$ and $\boldsymbol{K}_q^\mathcal{P} = \mathrm{diag}(90, 250, 60, 50, 5, 5, 1.2)$, $\boldsymbol{K}_q^\mathcal{D} = \mathrm{diag}(2, 4, 1, 0.5, 0.1, 0.1, 0.05)$. One demonstration was collected for each sub-task, as shown in Fig. 8, in different regions of the workspace. Here, variance is a measure of prediction uncertainty, unlike the previous task where it encoded variability, thus one demonstration is sufficient. For each output, we used a Gaussian Process with a Matérn kernel with $\nu = 3/2$ (see e.g., Chapter 4 in [15]), as it yielded smooth predictions, a convenient feature for our setup where the person may move in an unpredictable manner. Hyperparameters were optimized by minimizing the negative log marginal likelihood of the observations [15]. Moreover, we exploit the process mean $\boldsymbol{m}(\boldsymbol{x}^H)$ to define a prior on the robot's behavior, in particular to have the robot keep a safe posture outside of the region where demonstrations are provided. We define this neutral pose manually as a joint space configuration $\boldsymbol{m}_q = [0\ 0\ 0\ 1.1\ -0.2\ 0\ 0]^\top$ but it could alternatively be demonstrated. Each element $m_1^q, \ldots, m_7^q$ defines the mean of each of the 7 joint space GPs. The means of the task space GPs $\boldsymbol{m}_x$, also constant, are given by the end-effector position yielded by $\boldsymbol{m}_q$.

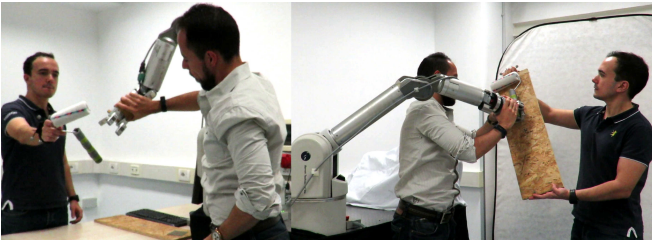

Fig. 8: Two persons demonstrate the painting task to the robot. **Left:** The robot is shown how to receive the paint roller. **Right:** One person drives the robot to demonstrate the painting strokes, while the other holds the board.

After hyperparameter estimation, we exploit GPR predictions to fuse the torques from each controller and reproduce the complete task. Notice that, during movement synthesis, the system will observe different input data than that used for training, as the user may move in regions where demonstrations were not provided. One expects the robot to stay in the pre-defined safe posture in those regions and execute the demonstrated sub-tasks where they were shown. Moreover, this should occur with smooth transitions between torque commands when tasks change. Figure 9 shows one reproduction of the complete task. The user starts by filing a wooden board, in a region of the workspace with no demonstration data (top, left). One can see that the robot remains in the pre-selected neutral pose. As the user hands the paint roller to the robot, the end-effector moves to grasp it (top, right). Finally, the user grasps the board and moves to a spacious region to perform the painting. As his right hand moves up and down, the robot applies painting strokes in the opposite direction. The robot is therefore capable of identifying which controller should be active at any moment, by exploiting the information contained in the data.

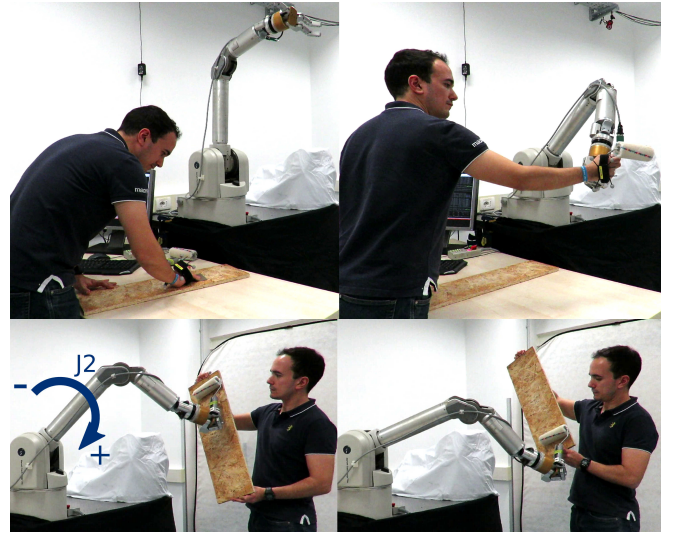

Fig. 9: Reproduction of the painting task. **Top:** The user works on a wooden board, while the robot keeps a safe posture (left). The paint roller is handed over to the robot (right). **Bottom:** The robot applies painting strokes, as the user's right hand moves up and down with the board.

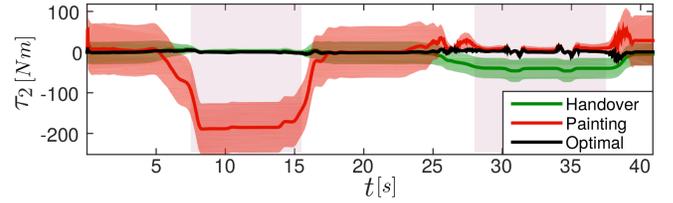

Fig. 10: Torques from the 2nd joint during the painting task and their variance. The first shaded area highlights the handover part of the movement, where the optimal torques match those computed by the end-effector position controller. The second shaded area highlights the task torques during two painting strokes.

Figure 10 provides a quantitative analysis of the performance of our method in this scenario, by showing the torques involved in one reproduction. We focus the analysis on the second joint of the robot (see Fig. 9, bottom left) since it is highly important for this task. Even though we did not consider a time-driven regression, we plot torques against time, in order to have a clear and continuous view of how the task evolved. The plot in Fig. 10 shows a clear separation between different moments of the task. Time intervals $0 - 7.5s$, $15 - 27.5s$, $37.5 - 40s$, correspond to regions of the workspace where no training data was provided and, thus, the variance of both controllers is high and roughly constant, as predictions are simultaneously uncertain. The interval $7.5 - 15s$ (first highlighted region) corresponds to the execution of the handover sub-task. Notice the decrease in the variance of the torques for this task (green envelope) and how these torques are matched by the optimal torque. Finally, the second highlighted time frame $27.5 - 37.5s$ coincides with the execution of the painting task. Here one can see a decrease in the variance of the joint space controller (red envelope), which is closely matched by the optimal torque, in particular during the two strokes (two oscillations around $30s$ and $35s$). All other joints yielded equivalent observations.

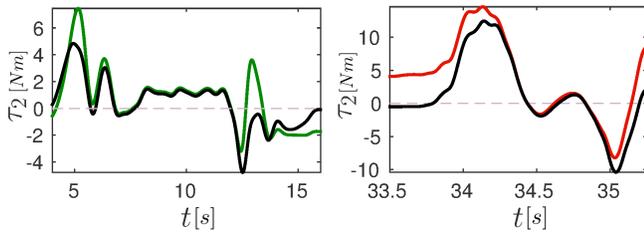

Fig. 11: Close up view of the handover and painting torques. **Left:** Optimal torque (black) and operational space controller torque (green). **Right:** Optimal and joint space controller torques (black and red).

For visualization purposes, in Fig. 11 we zoom in on the torques that are used for each sub-task. In the leftmost plot we see that the torques that are generated by the task space controller (green line) are closely matched by the optimal torque. Here, positive torques lower the end-effector to a below posture for the handover (until $t \approx 7.5s$), while negative torques raise it to an above posture after the handover ($t > 12s$). We observe an analogous result in the rightmost plot, where the joint space controller torques coincide. Here, positive torques apply vertical strokes from top to bottom, and negative torques move the paint roller back to the initial configuration.

## VII. CONCLUSIONS AND FUTURE WORK

We presented a novel probabilistic framework for fusing torque controllers based on human demonstrations. It improves on previous work by considering force-based, Cartesian position and joint space constraints as well as by being compatible with different probabilistic trajectory modeling techniques. The experimental validation showed that the approach allows robots to successfully reproduce manipulation tasks that require the fulfillment of different types of constraints, which are enforced by controllers acting on different spaces. The results presented here open up several future research challenges. One, connected to Section V, concerns the formulation of a probabilistic technique that can simultaneously encode and synthesize uncertainty and variability in the observed data. Works like [21] are a potential first step in this direction. Another promising research direction pertains to the design of the individual controllers. While in this paper we fixed the control gains, works like [10], [13], [22] estimate them from demonstrations by formulating the tracking problem as a LQR, which could allow us to alleviate the need for gain design and enhance safety in our framework. Finally, by exploiting the null space of the robot as in [23], we can possibly improve the extrapolation capabilities of the approach.